\def\year{2021}\relax

\documentclass[letterpaper]{article} 
\usepackage{aaai21}  
\usepackage{times}  
\usepackage{helvet} 
\usepackage{courier}  
\usepackage[hyphens]{url}  
\usepackage{graphicx} 
\usepackage{natbib}  
\usepackage{caption} 
\usepackage{booktabs}
\usepackage{array}
\title{A Free Lunch for Unsupervised Domain Adaptive Object Detection \\
without Source Data}
\author {Xianfeng Li, \textsuperscript{\rm 1}\thanks{Internship at Hikvision Research Institute.}
Weijie Chen, \textsuperscript{\rm 3}\textsuperscript{\rm 2}
Di Xie, \textsuperscript{\rm 2}
Shicai Yang, \textsuperscript{\rm 2}\\
Peng Yuan, \textsuperscript{\rm 2}
Shiliang Pu, \textsuperscript{\rm 2}\thanks{Shiliang Pu is the Corresponding Author}
Yueting Zhuang\textsuperscript{\rm 3}\\
}
\affiliations {
    \textsuperscript{\rm 1} South China University of Technology
    \textsuperscript{\rm 2} Hikvision Research Institute
    \textsuperscript{\rm 3} Zhejiang University \\
    lockonlxf@163.com, \{chenweijie5, xiedi, yangshicai, yuanpeng7, pushiliang.hri\}@hikvision.com, yzhuang@zju.edu.cn
}

\begin{document}

\maketitle

\begin{abstract}
\emph{Unsupervised domain adaptation} (UDA) assumes that source and target domain data are freely available and usually trained together to reduce the domain gap. However, considering the data privacy and the inefficiency of data transmission, it is impractical in real scenarios. Hence, it draws our eyes to optimize the network in the target domain without accessing labeled source data. To explore this direction in object detection, for the first time, we propose a \textit{source data-free domain adaptive object detection} (SFOD) framework via modeling it into a problem of learning with noisy labels. Generally, a straightforward method is to leverage the pre-trained network from the source domain to generate the pseudo labels for target domain optimization. However, it is difficult to evaluate the quality of pseudo labels since no labels are available in target domain. In this paper, \emph{self-entropy descent} (SED) is a metric proposed to search an appropriate confidence threshold for reliable pseudo label generation without using any handcrafted labels. Nonetheless, completely clean labels are still unattainable. After a thorough experimental analysis, false negatives are found to dominate in the generated noisy labels. Undoubtedly, false negatives mining is helpful for performance improvement, and we ease it to false negatives simulation through data augmentation like Mosaic. Extensive experiments conducted in four representative adaptation tasks have demonstrated that the proposed framework can easily achieve state-of-the-art performance. From another view, it also reminds the UDA community that the labeled source data are not fully exploited in the existing methods.
\end{abstract}

\begin{figure}[t]
	\centering
	\includegraphics[width=0.9\columnwidth]{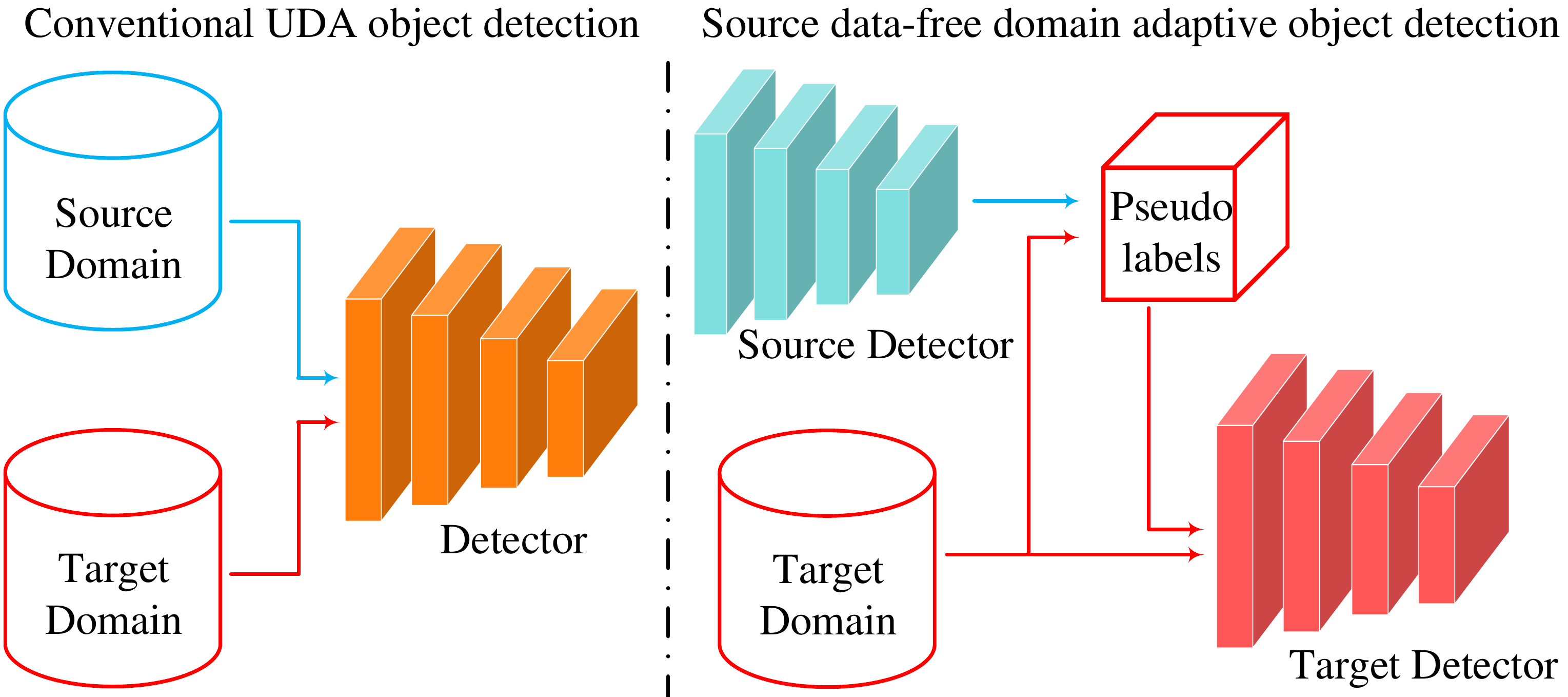} 
	\caption{The comparison between conventional UDA object detection and our proposed source data-free domain adaptive object detection.}
	\label{outline}
\end{figure}

\section{Introduction}
Deep convolutional neural networks have significantly improved object detection performance~\cite{Ren01,Redmon12,Liu13} but rely on large quantities of high-quality manual annotated training data. This limits the ability to generalize when facing new environments or data distributions where the object appearance, scene type, illumination, background, or weather condition are various. It attracts us to study how to transfer the pre-trained model from a label-rich source domain to an unlabeled target domain without supervision.

\begin{figure*}[t]
	\centering
	\includegraphics[width=1.65\columnwidth]{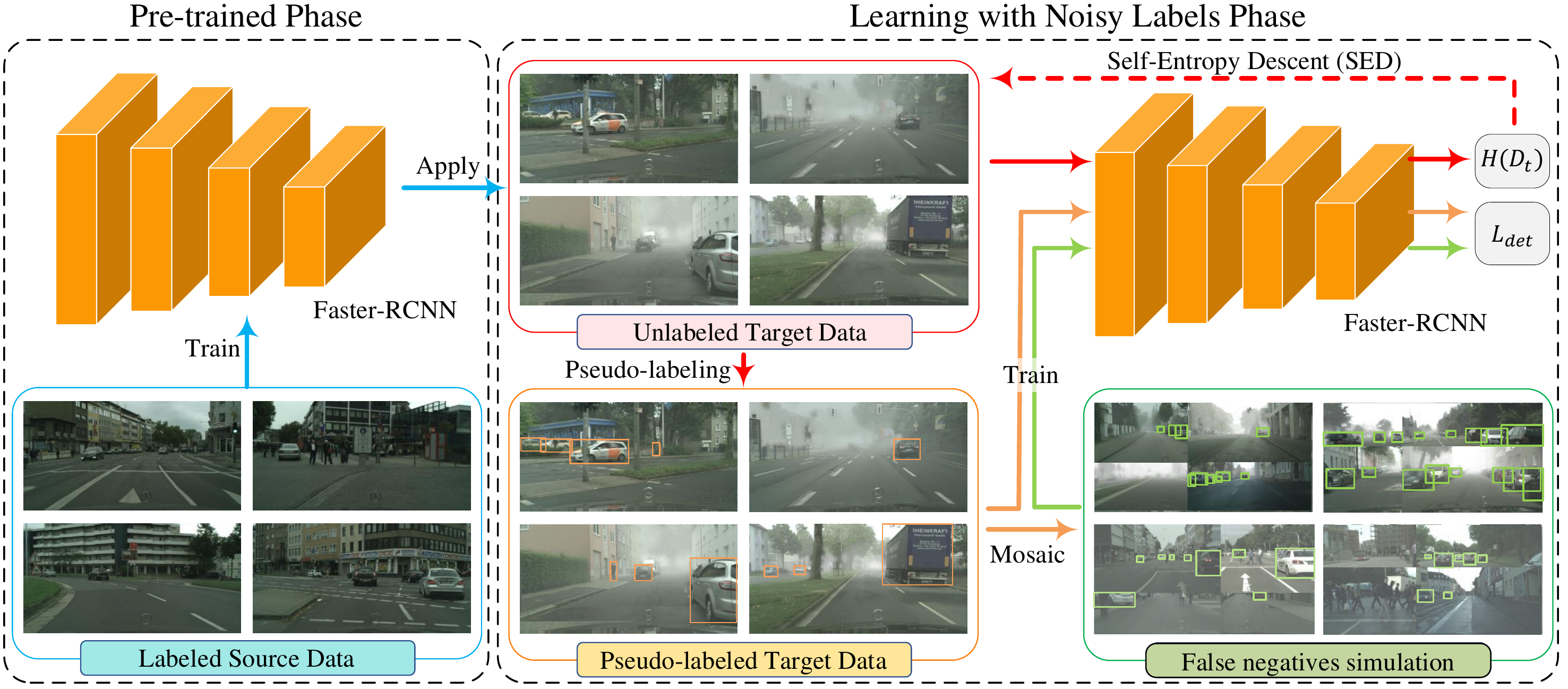} 
	\caption{The pipeline of the proposed source data-free domain adaptative object detection (SFOD). The given supervision signals are only provided by the pre-trained model from source domain during adaptation.}
	\label{pipeline}
\end{figure*}

Various unsupervised domain adaptive methods had been proposed to tackle this problem, whether using domain-invariant features for alignment \cite{Chen02,Saito03,He05,Xu06}, or narrowing the distribution of the domain in the image space~\cite{Liu08,Hoffman07,Hsu04}, or using pseudo label techniques by measuring the similarity between source and target domain samples~\cite{PFA2019,spbspl2020}. These methods align the distributions of source and target domain, assuming that the data distribution of labeled source domain and unlabeled target domain is related but different~\cite{Sugiyama09} and needing access freely to both source and target domain samples. However, this assumption will encounter challenges in practical application, such as data privacy and impractical data transmission.

At present, some classification methods \cite{Li11,Kim15,Peng20} about source data-free have made good progress, but there is still a blank in the source data-free unsupervised domain adaptive object detection. This paper proposes a simple yet effective approach to the above problems named \textit{source data-free domain adaptive object detection} (SFOD), which decouples the domain adaptation process by leveraging a pre-trained source model. The key idea is to train the target model with reliable pseudo labels of target samples in a self-learning manner. The natural question is how to evaluate the quality of pseudo labels for object detection and learn with noisy labels. In the classification task, the total number of samples is fixed. Even if only a few reliable pseudo labels are used, more reliable pseudo labels can be mined to achieve pseudo labels refinement~\cite{Kim15}. However, it is more challenging in the object detection task, since the negative samples are countless and various, and lots of hard positive samples are difficult to box out and mixed with negative samples. Only relying on a small number of reliable samples cannot achieve good performance. A straightforward method is to directly filter out the bounding boxes into positive parts and negative parts according to an appropriate confidence threshold. Although there unavoidably are some false positive and false negative samples (namely noisy labels in object detection), the target model can still be optimized following the 'wisdom of the crowd' principle. However, an appropriate confidence threshold is difficult to search since no metric is available for supervision. It would harm the network performance if confidence threshold is too high or too low due to the messier noisy labels. This inspires us to search for an appropriate confidence threshold for pseudo label generation to make a trade-off between the positive effect brought by true positives and the negative effect brought by false positives and false negatives.

In this paper, a metric named \textit{self-entropy descent} (SED) is proposed to search the confidence threshold. As is known, prediction uncertainty can be quantified as self-entropy, i.e., $H(x)=-\sum p(x) \log (p(x))$. The lower the self-entropy the more confident the prediction. Here we search the confidence threshold from the higher score to the lower score. Meanwhile, we use the generated pseudo label to fine-tune the pretrained model and then evaluate the self-entropy of the dataset after training (namely mean self-entropy). Note that the noisier the labels, the more difficult to fit the labels. Therefore, as the confidence threshold decreases, when the mean self-entropy descends and hits the first local minimum, we select it as an appropriate confidence threshold for reliable pseudo label generation. We design a toy experiment to prove the reasonability of this solution. Nonetheless, we have to admit the generated pseudo labels are still unavoidably noisy. Specifically, there exist false positives and false negatives in the generated pseudo labels. Through a thorough experimental analysis in the publicly-released datasets, false negatives are found to dominate in the noisy labels, such as small and obscured objects. Hence, to alleviate the effects from false negatives, false negatives mining is proposed to solve this problem. And we ease this solution to false negatives simulation via data augmentation like Mosaic \cite{Bochkovskiy43}, since it can exploit the easy positive samples to simulate false negative samples. We believe more label denoising techniques can further boost the performance, and we leave this as our future work. 

The main contributions of this work are summarized as follows. (i) To the best of our knowledge, this is the first work on source data-free unsupervised domain adaptative object detection. (ii) We innovatively model the source data-free UDA into a problem of learning with noisy labels and make it solvable. (iii) Our framework can achieve delectable performance without using source data and surpass most of the other source data based UDA methods. It implies the UDA community that the labeled source data are not fully exploited in the existing UDA methods in object detection.

\section{Related Works}

\subsection{Domain Adaptive Object Detection}
The proposal of \textit{Domain Adaptive Faster R-CNN}~\cite{Chen02} has made progress in the challenging unsupervised domain adaptative object detection task, which aligns both the image and instance levels in a domain adversarial manner. After that, the following works \textit{Strong-Weak Domain Adaptive Faster R-CNN}~\cite{Saito03}, \textit{Region-level Alignment}~\cite{Zhu16}, \textit{Categorical Regularization Domain Adaptive Object Detection}~\cite{Xu06}, \textit{Asymmetric Tri-way Faster-RCNN}~\cite{He05}, and style transfer based method~\cite{Hsu04} were proposed one after another to push this direction forward. However, the existing methods require both labeled source data and unlabeled target data, while our proposed source data-free one is more practical in real scenarios.

\begin{figure}[t]
	\centering
	\includegraphics[width=0.9\columnwidth]{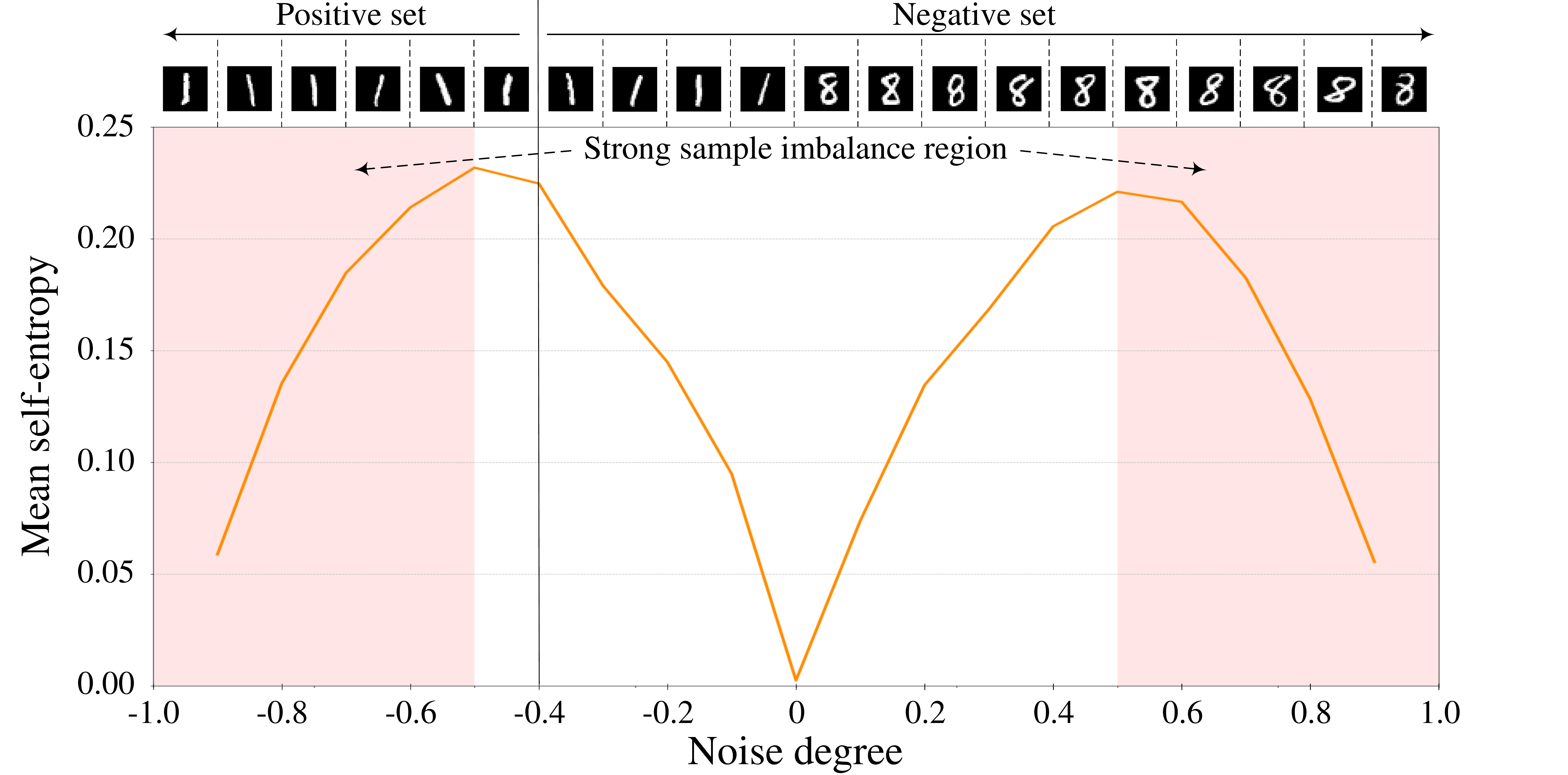} 
	\caption{A toy example to capture the relation between noisy labels and mean self-entropy. Noise degree denotes the ratio of positive samples mixed into negative set (-) and the ratio of negative samples mixed into positive set (+). Two local minimums appearing in two ends of the curve are resulted from strong sample imbalance.}
	\label{toy_example}
\end{figure}
\begin{figure}[t]
	\centering
	\includegraphics[width=0.8\columnwidth]{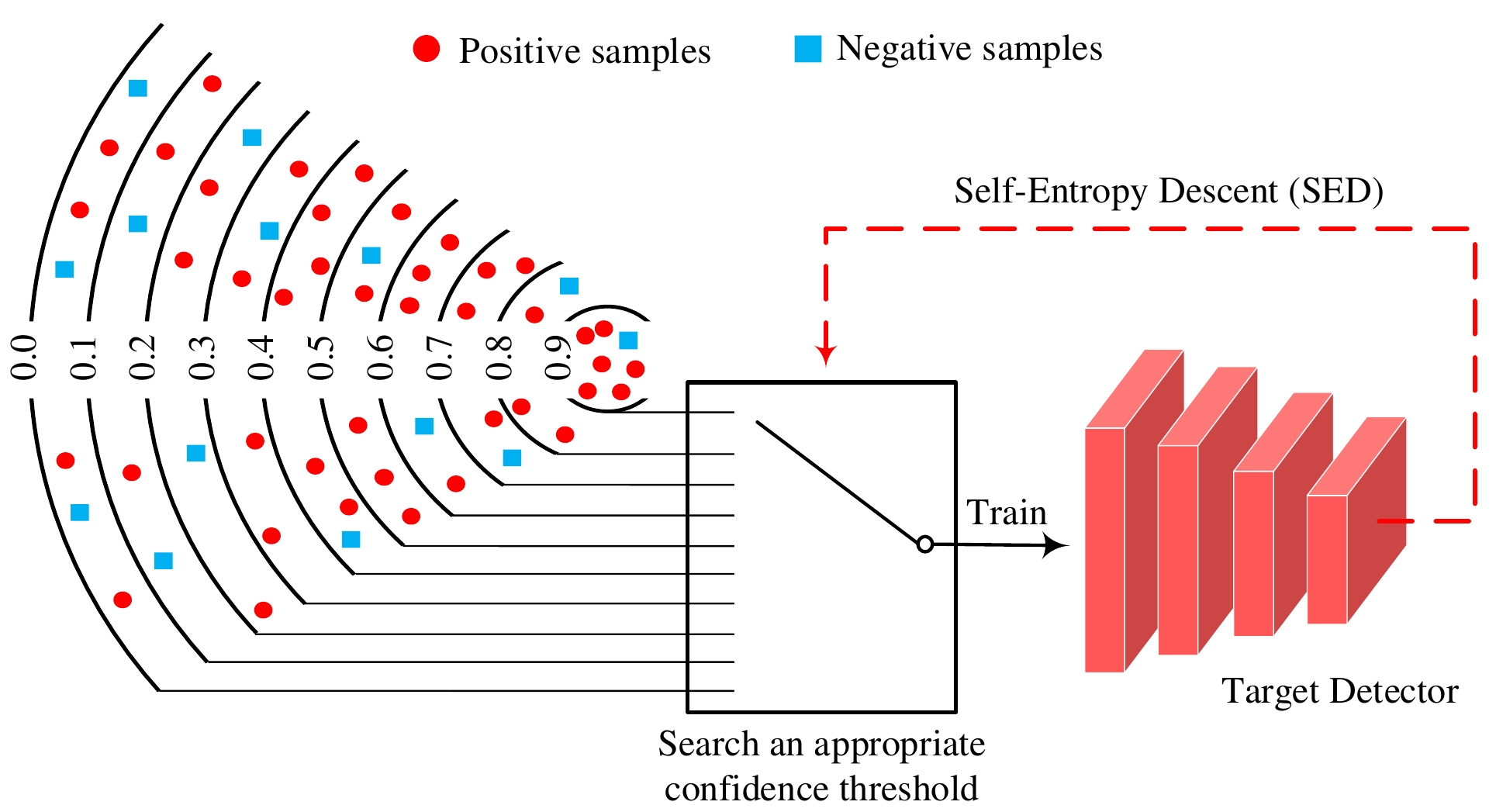} 
	\caption{An appropriate confidence threshold to split positive and negative objects is searched from the higher score to the lower score with the metric of SED.}
	\label{sed}
\end{figure}

\subsection{Domain Adaptation without Source Data}
Considering data privacy and data transmission, some source data-free domain adaptative classification approaches have been proposed ~\cite{Li11,Kim15}. 
However, there is still a blank in source data-free unsupervised domain adaptive object detection.

\subsection{Learning with Noisy Labels}
The current research about learning with noisy labels still focuses on relatively simple classification tasks. Earlier work in this field used an instance-independent noise model, where each class was confused with other classes that were independent of the content of the instance \cite{Mnih58,Natarajan54,Patrini55}. Recently, some methods focus on specific examples of label noise prediction \cite{Vahdat56,Veit57,Ren52,Jiang53}. However, the noisy labels setting in these researches is ideal, where the noisy labels and true labels are manually set and identically distributed. While in our work, the noisy labels are not manually set and not identically distributed, where the noisy labels are both hard positive and negative objects. Moreover, there are few methods about learning with noisy labels designed for object detection. ~\cite{Khodabandeh17} is the one easing the object detection problem into image classification, but it cannot solve the situation when the objects are hard to box out.

\section{Source free Domain Adaptive Object Detection}
The assumption of unsupervised domain adaptive object detection is that the data of labeled source domain $D_{s}=\left\{\left(x_{s}^{i}, y_{s}^{i}\right)\right\}_{i=1}^{N_{s}}$ and unlabeled target domain $D_{t}=\left\{x_{t}^{i}\right\}_{i=1}^{N_{t}}$ are available freely in training step to minimize the discrepancy between them. Unlike this learning paradigm, source data-free UDA aims to optimize the network only through the unlabeled target domain $D_{t}=\left\{x_{t}^{i}\right\}_{i=1}^{N_{t}}$. The only supervision signal is given by the pre-trained model $\theta_{s}$ from source domain instead of directly using source domain data.

\subsection{Pseudo Labels Optimization via SED}
\subsubsection{A toy example: How to evaluate the quality of pseudo labels?}
In this section, a toy example on two categories of MNIST~\cite{LeCun41} dataset representing the positive and negative samples, called MNIST-2, is presented. To study how to evaluate the quality of pseudo labels, we build different datasets based on MNIST-2 through mixing different proportions of positive samples into the negative part or mixing different proportions of negative samples into the positive part and use LeNet~\cite{Lecun42} to train these datasets. For simplicity, the mixing proportion is also named as noise degree. And a notion named mean self-entropy is introduced to capture the uncertainty of the prediction of the entire dataset after training which can be formulated as follows:
\begin{equation}
H\left(D_{t}\right)=-\frac{1}{N_{t}} \sum_{i}^{N_{t}}\left(\frac{1}{n_{c}} \sum_{c}^{n_{c}} p_{c}\left(x_{t}^{i}\right) \log \left(p_{c}\left(x_{t}^{i}\right)\right)\right)
\end{equation}
where $n_{c}$ refer to the class number, and $p_{c}\left(x_{t}^{i}\right)$ denotes the prediction probability of class $c$, respectively.

Unsurprisingly, as shown in Figure \ref{toy_example}, the noise degree is positively correlated with mean self-entropy. The noisier the labels, the more difficult to fit the labels, which leads to larger mean self-entropy. Note that two local minimums in two ends of the mean self-entropy curve are resulted from a strong sample imbalance. Ideally, the cleanest label assignment will lead to the lowest mean self-entropy. Considering both situations, it indicates a reliable label assignment when mean self-entropy descends and hits the local minimum.

\begin{figure}[t]
	\centering
	\includegraphics[width=0.9\columnwidth]{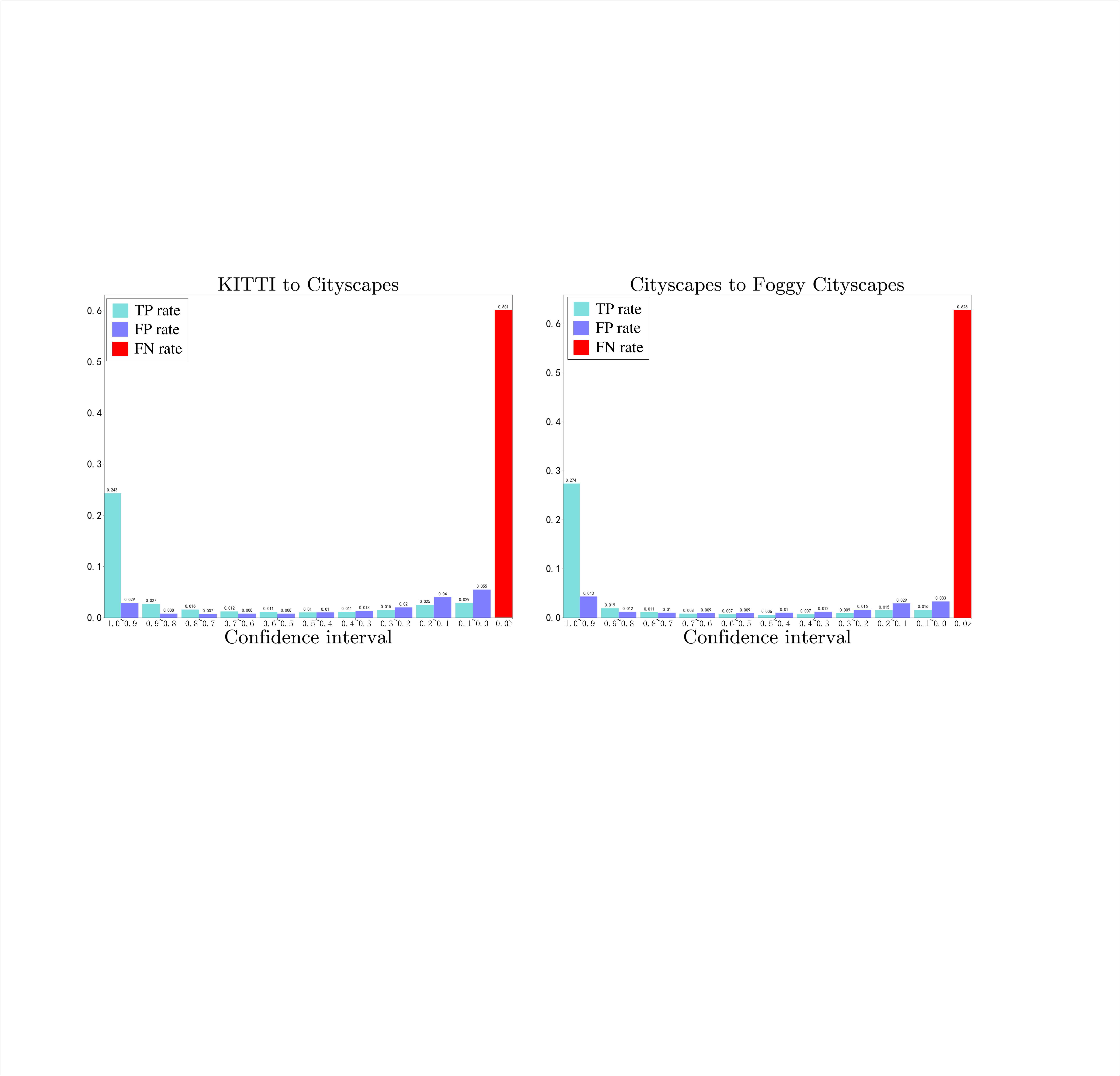} 
	\caption{In two cross-domain datasets, KITTI to Cityscapes and Cityscapes to Foggy Cityscapes, the ratio of true positives and false positives to the entire ground truth are counted in different confidence intervals. The confidence of target domain data is directly predicted by the pre-trained model from source domain. False negatives ($<$0.0), which are difficult to box out even when the confidence threshold is set to zero, are found to dominate in the noisy labels.}
	\label{recall}
\end{figure}
\begin{figure}[t]
	\centering
	\includegraphics[width=0.9\columnwidth]{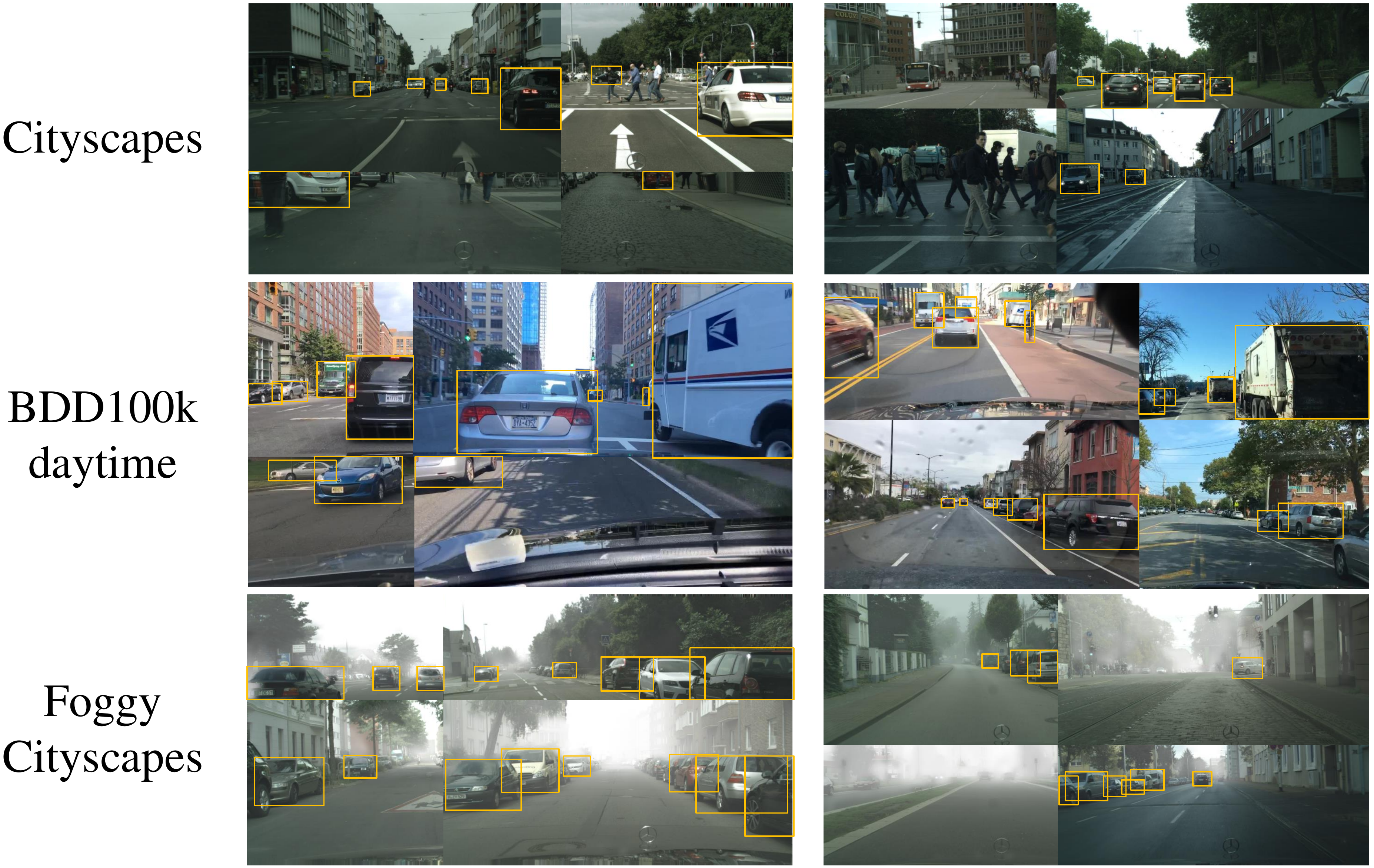} 
	\caption{Mosaic visualization with pseudo labels.}
	\label{mosaic}
\end{figure}

\subsubsection{Self-entropy descent: how to generate reliable pseudo labels in object detection?}

When it comes to object detection, the negative samples are countless and various. Based on the clue in the above section, we search an appropriate confidence threshold from the higher score to the lower score to split positive and negative samples for training and stop when the mean self-entropy descends and hits the FIRST local minimum. We name it as \textit{Self-entropy descent}. 

The unlabeled target domain $D_{t}=\left\{x_{t}^{i}\right\}_{i=1}^{N_{t}}$ and the source domain's pre-trained model $\theta_{s}$ are available freely. So pseudo labels $y\left(x_{t}\right)$ and the corresponding confidence $p\left(x_{t}\right)$ can be obtained as follows:
\begin{equation}\left\{y\left(x_{t}^{i}\right), p\left(x_{t}^{i}\right)\right\}_{i=1}^{N_{t}}=\left\{F\left(x_{t}^{i} \mid h, \theta_{s}\right)\right\}_{i=1}^{N_{t}}\end{equation}
where $h$ is a confidence threshold for pseudo label generation, and $F$ represents Faster-RCNN~\cite{Ren01} detector. Faster-RCNN is the first anchor-based object detection method, where the detector has an encoder network as a feature extractor, a \textit{Region Proposal Network} (RPN) and \textit{Region of Interest} (ROI) classifier. 

Specifically, the confidence $p\left(x_{t}\right)$ is the output of softmax in the classification branch. And the pseudo label $y\left(x_{t}\right)$ is determined by the argmax of foreground class probability. If this score is greater than the given confidence threshold $h$, the corresponding box will be assigned as the class label with the max score; otherwise, it will be assigned as the background class. To train the target domain data with pseudo labels, the loss function is formulated as:
\begin{equation}
	L_{det}=L_{rpn}+L_{cls}+L_{reg}
\end{equation}
where $L_{rpn}$, $L_{cls}$, and $L_{reg}$ denotes the region proposal loss, region classification loss and the bounding-box regression loss. As for region proposal and bounding boxes regression, we directly use the bounding boxes predicted by the pre-trained model from the source domain as ground-truth for training. As claimed by \cite{borji2019empirical}, the location error is much weaker than classification error in object detection task. 

After fine-tuning the pre-trained model with the pseudo-labels generated by a given confidence threshold, we use the updated model $\theta_{t}$ to evaluate the mean self-entropy $H\left(D_{t}\right)$ of the target datasets.
\begin{equation}
\theta_{t}=Train(\{x_{t}^{i}, y_{t}^{i}\} | \theta_{s})
\end{equation}
\begin{equation}
\left\{-,\hat{p}\left(x_{t}^{i}\right)\right\}_{i=1}^{N_{t}}=\left\{F\left(x_{t}^{i} \mid h, \theta_{t}\right)\right\}_{i=1}^{N_{t}}
\end{equation}
\begin{equation}
H\left(D_{t}\right)=-\frac{1}{N_{t}} \sum_{i}^{N_{t}}\left(\frac{1}{n_{c}} \sum_{c}^{n_{c}} \hat{p}_{c}\left(x_{t}^{i}\right) \log \left(\hat{p}_{c}\left(x_{t}^{i}\right)\right)\right)
\end{equation}
According to the SED policy, we search the confidence threshold from the higher to the lower score, and early stop when $H\left(D_{t}\right)$ descends and hit the first local minimum.
\begin{equation}
h_{optimal}=\mathop{\arg\min}_{h} H\left(D_{t}\right)
\end{equation}

\subsection{False Negatives Simulation}
\begin{figure*}[t]
	\centering
	\includegraphics[width=0.83\textwidth]{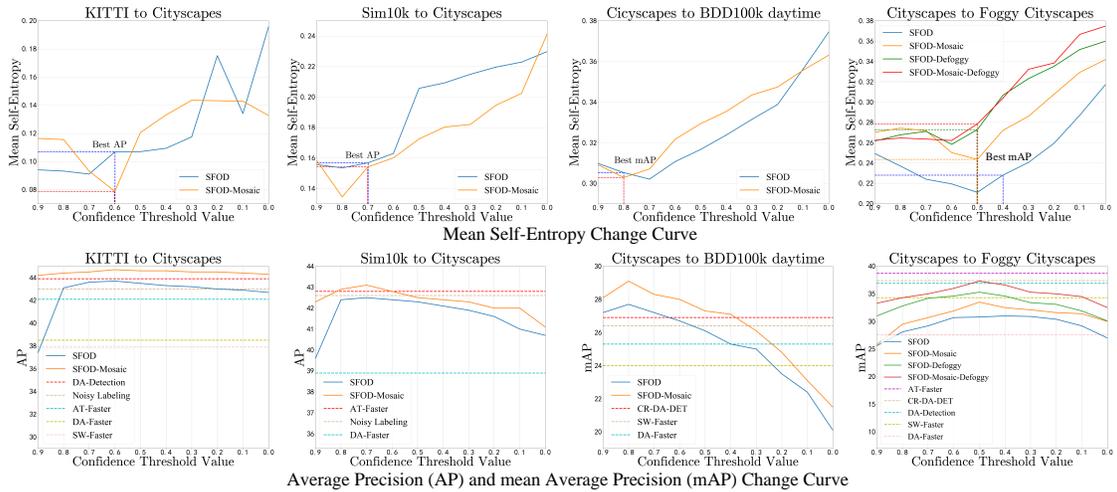} 
	\caption{The curves of mean self-entropy and the corresponding AP or mAP vary with confidence threshold in four adaptation tasks. It can nearly search the best mAP via SED.}
	\label{curve4}
\end{figure*}

Although we search for an appropriate confidence threshold via SED, we have to admit the generated pseudo labels are still noisy. Label denoise techniques can be applied to clean the labels and boost performance. In an object detection task, the noisy labels behave as false positives and false negatives. We count the true positives and false positives in each confidence interval in several publicly-released datasets. As shown in Figure \ref{recall}, false positives only account for a relatively small proportion. And surprisingly, more than 50\% positive samples are difficult to box out even though we set the confidence threshold close to zero, which behave as false negatives during training. Therefore, in this paper, we mainly focus on false negatives mining for labels denoising.

Through visualization, most false negatives are small and obscured objects mixed with true negatives, which are very difficult to mine back into the positive part. The domain gap between the source domain and target domain increases the difficulty of detecting hard examples. Hence, we ease this solution to false negatives simulation by exploiting true positives. Data augmentation is a good way to augment the detected positives into hard ones to simulate the small and obscured objects. It can suppress the negative effects of false negatives. In this work, Mosaic augmentation \cite{Bochkovskiy43} is selected for false negatives simulation since it can generate small-scale and blocked objects by exploiting true positives while not harming the true negatives. Mosaic is the improvement of CutMix \cite{Yun44} via mixing four training images, which allows the detection of objects outside their normal context. The two main steps in Mosaic are random scaling and random cutting. The hard objects with different scales can be simulated by using the simple objects that have been detected in the target domain via using random scaling. Meanwhile, the blocked objects with the only visible part of the structure can be simulated to a certain extent by random cutting. Mosaic data $\{\left(\tilde{x},\tilde{y}\right)\}$ can be formulated by the target domain data $\{\left(x_{A},y_{A}\right),\left(x_{B},y_{B}\right),\left(x_{C},y_{C}\right),\left(x_{D},y_{D}\right)\}$ as follows:

\begin{table}[]
	\centering
	\begin{tabular}{l|c}
		\specialrule{0.075em}{0pt}{1pt}
		Methods & AP of Car \\ 
		\specialrule{0.05em}{1pt}{1pt}
		Source only         & 36.4      \\ \specialrule{0.05em}{1pt}{1pt}
		DA-Faster~\cite{Chen02}            & 38.5      \\ 
		SW-Faster~\cite{Saito03}          & 37.9      \\ 
		MAF~\cite{2019Multi} & 41.0 \\
		AT-Faster~\cite{He05}          & 42.1      \\ 
		Noise Labeling~\cite{Khodabandeh17}    & 43.0      \\ \specialrule{0.05em}{1pt}{1pt}
		DA-Detection~\cite{Hsu04}       & 43.9      \\ 
		SFOD (SED)               & 43.6      \\ 
		SFOD-Mosaic (SED)        & 44.6   \\ 
		SFOD (Ideal)        & 43.7      \\ 
		SFOD-Mosaic (Ideal) & 44.6      \\ \specialrule{0.05em}{1pt}{1pt}
		Oracle              & 58.5      \\ \specialrule{0.075em}{1pt}{0pt}
	\end{tabular}
	\caption{Results of adaptation to a new sense, i.e., from KITTI dataset to Cityscapes dataset.}
	\label{K2C}
\end{table}

\begin{equation}\tilde{x}=\left[\begin{array}{ll}
M_{A} \odot s\left(x_{A}\right) & M_{B} \odot s\left(x_{B}\right) \\
M_{C} \odot s\left(x_{C}\right) & M_{D} \odot s\left(x_{D}\right)
\end{array}\right] \in R^{W \times H}\end{equation}

\begin{equation}\begin{array}{l}
\tilde{y}=(\lambda, \lambda) \cdot s\left(y_{A}\right)+(\lambda, 1-\lambda) \cdot\left(s\left(y_{B}\right)+v\right) \\
\qquad+(1-\lambda, \lambda) \cdot\left(s\left(y_{c}\right)+u\right)\\
\qquad+(1-\lambda, 1-\lambda) \cdot\left(s\left(y_{D}\right)+(u, v)\right)
\end{array}\end{equation}
where $W$ and $H$ represent the size of training images, $\left\{M_{A}, M_{B}, M_{C}, M_{D}\right\} \in\{0,1\}^{s\left(W\right) \times s\left(H\right)}$ denotes a group of binary masks, $\left(u,v\right)$ is the 2D translation,  $s\left(\cdot\right)$ and $\lambda$ represent random scaling function and random cutting factor. Figure \ref{mosaic} displays some Mosaic images. We believe more effective false negatives mining or false negatives simulation methods can bring further performance boost.

False negatives simulation is adopted with SED to search an appropriate confidence threshold for pseudo label generation. The entire pipeline of SFOD is shown in Figure \ref{pipeline}.

\begin{table}[]
	\centering
	\begin{tabular}{l|c}
		\specialrule{0.075em}{0pt}{1pt}
		Methods & AP of Car \\ \specialrule{0.05em}{1pt}{1pt}
		Source only         & 33.7      \\ \specialrule{0.05em}{1pt}{1pt}
		DA-Faster~\cite{Chen02}         & 38.5      \\ 
	    MAF~\cite{2019Multi} & 41.1 \\
		AT-Faster~\cite{He05}         & 42.1      \\ 
		Noise Labelling~\cite{Khodabandeh17}    & 43.0      \\ \specialrule{0.05em}{1pt}{1pt}
		SFOD (SED)               & 42.3      \\ 
		SFOD-Mosaic (SED)        & 42.9      \\ 
		SFOD (Ideal)      & 42.5      \\ 
		SFOD-Mosaic (Ideal) & 43.1      \\ \specialrule{0.05em}{1pt}{1pt}
		Oracle              & 58.5      \\ \specialrule{0.075em}{1pt}{0pt}
	\end{tabular}
	\caption{Results of adaptation from synthetic to real images, i.e.,from Sim10k dataset to Cityscapes dataset.}
	\label{S2C}
\end{table}

\section{Experiments}

\begin{table*}[]
	\centering
	\begin{tabular}{l|cccccccc|c}
		\specialrule{0.075em}{0pt}{1pt}
		Methods & truck & car  & rider & person & train & motor & bicycle & bus  & mAP  \\ \specialrule{0.05em}{1pt}{1pt}
		Source only          & 14.0  & 40.7 & 24.4  & 22.4   & -     & 14.5       & 20.5    & 16.1 & 21.8 \\ \specialrule{0.05em}{1pt}{1pt}
		DA-Faster~\cite{Chen02}            & 14.3  & 44.6 & 26.5  & 29.4   & -     & 15.8       & 20.6    & 16.8 & 24.0 \\ 
		SW-Faster~\cite{Saito03}            & 15.2  & 45.7 & 29.5  & 30.2   & -     & 17.1       & 21.2    & 18.4 & 25.3 \\ 
		CR-DA-DET~\cite{Xu06}    & 19.5  & 46.3 & 31.3  & 31.4   & -     & 17.3       & 23.8    & 18.9 & 26.9 \\ \specialrule{0.05em}{1pt}{1pt}
		SFOD (SED)                & 20.4  & 48.8 & 32.4  & 31.0   & -     & 15.0       & 24.3    & 21.3 & 27.6 \\ 
		SFOD-Mosaic (SED)         & 20.6  & 50.4 & 32.6  & 32.4   & -     & 18.9       & 25.0    & 23.4 & 29.0 \\
		SFOD (Ideal)       & 20.0  & 46.8 & 32.1  & 31.5   & -     & 16.3       & 25.1    & 21.8 & 27.7 \\ 
		SFOD-Mosaic (Ideal) & 20.6  & 50.4 & 32.6  & 32.4   & -     & 18.9       & 25.0    & 23.4 & 29.0 \\ \specialrule{0.05em}{1pt}{1pt}
		Oracle               & 53.4  & 53.5 & 42.8  &  41.9  & -     & 37.3	   & 38.8    & 58.1 & 47.1 \\ \specialrule{0.075em}{0pt}{1pt}
	\end{tabular}
	\caption{Results of adaptation to a large-scale dataset, i.e.,from Cityscapes dataset to BDD100k daytime dataset.}
	\label{C2B}
\end{table*}

\subsection{Experimental Setup}
\subsubsection{Datasets}
Five public datasets are utilized in our experiments. (1) \textbf{KITTI} \cite{Geiger45} is a popular dataset for autonomous driving, which are manually collected in several different scenes in a city with 7,481 labeled images for training. (2) \textbf{Sim10k} \cite{Johnson-Roberson46} simulates different scenes, such as different times or weather, from a computer game \textit{Grand Theft Auto V} (GTA V) with 10k images. (3) \textbf{Cityscapes} \cite{Cordts47} focuses on the high variability of outdoor street scenes from different cities. We transform the instance segmentation annotations of 2,975 training images and 500 validation images into bounding boxes for our experiments. (4) \textbf{BDD100k} \cite{Yu48} includes 100k images with six types of weather, six different scenes, and three categories for the time of day. We extract a subset labeled as daytime, including 36,728 training and 5,258 validation images. (5) \textbf{Foggy Cityscapes} \cite{Sakaridis49} simulates the foggy weather using city images from Cityscapes with three foggy weather levels and inherit annotations of Cityscapes.

\subsubsection{Implementation Details}
For a fair comparison with existing approaches, we follow the same experimental setting as \cite{Chen02,Xu06}. The short size of all training and testing images are resized to a length of 600 pixels. We use the pre-trained weights of VGG-16 \cite{Simonyan50} on ImageNet \cite{Deng51} as the backbone of the Faster-RCNN framework. The detector is trained with \textit{Stochastic Gradient Descent} (SGD) with a learning rate of 0.001. The batch size is set to 1. Source domain data are only used in the pre-trained step.
\subsection{Comparison Results}

\begin{figure*}[t]
	\centering
	\includegraphics[width=0.91\textwidth]{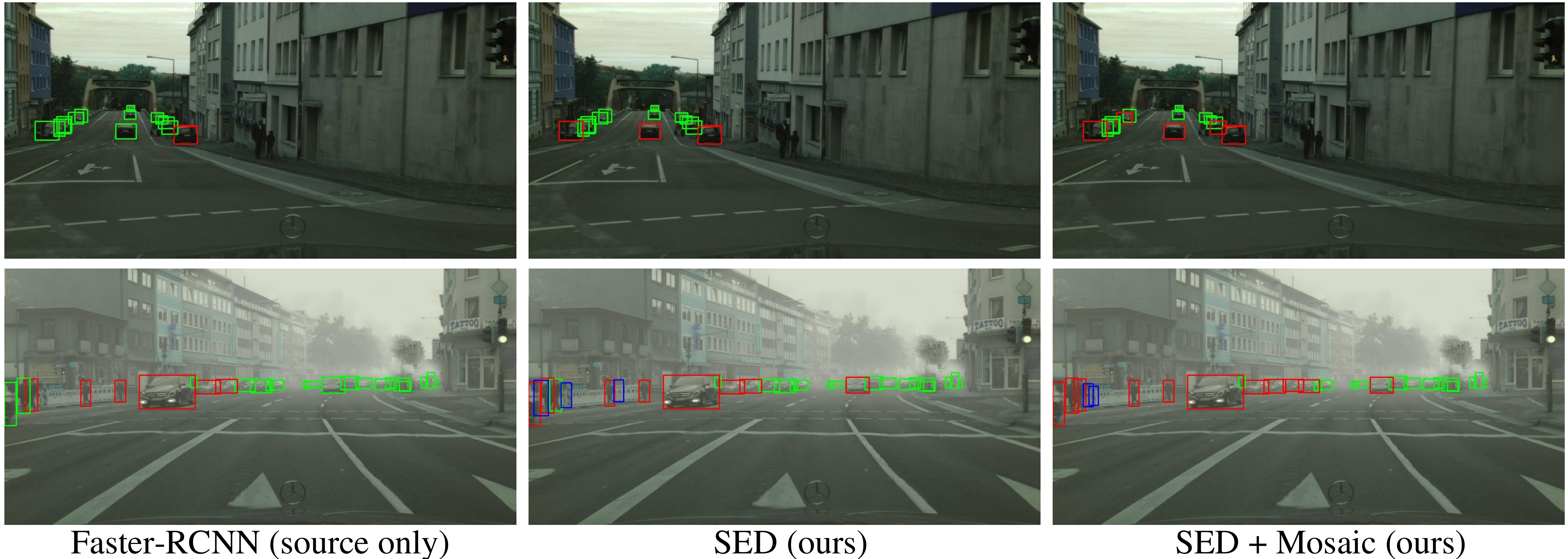} 
	\caption{Qualitative results. Top: KITTI to Cityscapes. Bottom: Cityscapes to Foggy Cityscapes. Red, green and blue boxes denote true positives, false negatives and false positives.}
	\label{result}
\end{figure*}
Our experiments are carried out in four adaptation tasks. Figure \ref{curve4} shows the curves of detection precision and mean self-entropy under different confidence thresholds for pseudo label generation. "Source only" and "Oracle" are both tested in target domain validation set, but trained with labeled source domain training set and labeled target domain training set, respectively.
\subsubsection{Adaptation to A New Sense}
Different camera setups (e.g., angle, resolution, quality, and type) widely exist in the real world, which can cause the domain shift. In this experiment, we take the adaptation to a new sense task between two real datasets. The KITTI and Cityscapes datasets are used as source and target domains, respectively. We implement our SFOD, DA-Faster~\cite{Chen02}, SW-Faster~\cite{Saito03}, Noise Labeling~\cite{Khodabandeh17}, DA-Detection~\cite{Hsu04}, and AT-Faster~\cite{He05} in this task. In Table \ref{K2C}, the \textit{average precision} (AP) on the car category, the only common object, is compared. When SED is used alone, although the ideal confidence threshold searched by the labeled target validation set is not found, the AP is very close to the ideal one, and our method has surpassed many existing methods in terms of car detection accuracy. When Mosaic is further used, the AP can be increased from 43.6\% to 44.6\% and exceeds DA-Detection~\cite{Hsu04} by 0.7\%. We can see that false negatives simulation can ease the negative effect brought by the false negative noisy labels.

\subsubsection{Adaptation from Synthetic to Real Images}
Another domain adaptation scenario is from synthetic data to the real world. Due to the lack of annotated training data to autonomous driving, synthetic data offers an alternative. Thus, the source domain is the Sim10k, and the target domain is the Cityscapes. In this task, we only evaluate the performance in annotated cars for which is the only object category in both Sim10k and Cityscapes. In Table \ref{S2C}, compared with DA-Faster~\cite{Chen02}, Noise Labeling~\cite{Khodabandeh17}, and AT-Faster~\cite{He05}, our source data-free method can achieve superior or comparable results. However, our source data-free setting is more challenging than the existing source data-based methods.

\begin{table*}[]
	\centering
	\begin{tabular}{l|c|cccccccc|c}
		\specialrule{0.075em}{0pt}{1pt}
		Methods & defoggy & truck & car  & rider & person & train & motor & bicycle & bus  & mAP  \\ \specialrule{0.05em}{1pt}{1pt}
		Source only                &$\times$ & 11.6  & 38.7 &31.4  &23.6   & 9.4   & 17.3       & 27.4     & 19.0     & 22.3     \\ \specialrule{0.05em}{1pt}{1pt}
		DA-Faster~\cite{Chen02}                    &$\times$ & 19.5  & 43.5 & 36.5  & 28.7   & 12.6  & 24.8       & 29.1    & 33.1 & 28.5 \\ 
		MAF~\cite{2019Multi} &$\times$ & 23.8 & 43.9 & 39.5 & 28.2 & 33.3 & 29.2 & 33.9 & 39.9 & 34.0 \\
		SW-Faster~\cite{Saito03}                 &$\times$ & 23.7  & 47.3 & 42.2  & 32.3   & 27.8  & 28.3       & 35.4    & 41.3 & 34.8 \\ 
		DA-Detection~\cite{Hsu04}               &$\surd$  & 24.3  & 54.4 & 45.5  & 36.0   & 25.8  & 29.1       & 35.9    & 44.1 & 36.9 \\ 
		CR-DA-DET~\cite{Xu06}         &$\times$ & 27.2  & 49.2 & 43.8  & 32.9   & 36.4  & 30.3       & 34.6    & 45.1 & 37.4 \\ 
		AT-Faster~\cite{He05}                 &$\times$ & 23.7  & 50.0 & 47.0  & 34.6   & 38.7  & 33.4       & 38.8    & 43.3 & 38.7 \\ \specialrule{0.05em}{1pt}{1pt}
		SFOD (SED)                        &$\times$ & 21.7  & 44.0 & 40.4  & 32.6   & 11.8  & 25.3       & 34.5    & 34.3 & 30.6 \\ 
		SFOD-Mosaic (SED)                 &$\times$ & 25.5  & 44.5 & 40.7  & 33.2   & 22.2  & 28.4       & 34.1    & 39.0 & 33.5 \\ 
		SFOD (Ideal)                &$\times$ & 22.3  & 44.0 & 38.2  & 31.4   & 15.1  & 25.7       & 34.6    & 36.8 & 31.0 \\ 
		SFOD-Mosaic (Ideal)         &$\times$ & 25.5  & 44.5 & 40.7  & 33.2   & 22.2  & 28.4       & 34.1    & 39.0 & 33.5 \\ \specialrule{0.05em}{1pt}{1pt}
		SFOD-Defoggy (SED)              & $\surd$ & 28.4  & 50.9 & 41.6  & 32.2   & 15.9  & 28.1       & 36.0    & 40.1 & 34.2 \\ 
		SFOD-Mosaic-Defoggy (SED)         & $\surd$ & 27.9  & 51.7 & 44.7  & 33.2   & 21.3  & 28.6       & 37.3    & 45.9 & 36.3 \\ 
		SFOD-Defoggy (Ideal)        & $\surd$ & 26.2  & 50.6 & 41.8  & 32.5   & 24.4  & 28.7       & 36.1    & 40.5 & 35.1 \\ 
		SFOD-Masoic-Defoggy (Ideal) & $\surd$ & 30.4  & 51.9 & 44.4  & 34.1   & 25.7  & 30.3       & 37.2    & 41.8 & 37.0 \\ \specialrule{0.05em}{1pt}{1pt}
		Oracle                       &$\times$ & 38.1  & 49.8 &53.1  &33.1   & 37.4  & 41.1       & 57.4     & 48.2     & 44.8 \\ \specialrule{0.075em}{1pt}{0pt}
	\end{tabular}
	\caption{Results of adaptation from normal to foggy dataset, i.e.,from Cityscapes dataset to Foggy Cityscapes dataset.}
	
	\label{C2F}
\end{table*} 

\subsubsection{Adaptation to Large-Scale Dataset }

Currently, collecting large amounts of image data is not difficult, but labeling those data is still the main problem for supervised learning methods. In this experiment, we use Cityscapes as a smaller source domain dataset, BDD100k containing distinct attributes as a large unlabeled target domain dataset. Since there is only daytime data in Cityscapes, we select the labeled daytime data in the three-time periods of BDD100k as the target domain. We evaluate the \textit{mean average precision} (mAP) of detection results on seven categories in both datasets. As we can see from the baseline and oracle results in Table \ref{C2B}, resolving such a domain divergence between a source domain and a target domain is so complicated that only a handful of approaches (e.g., DA-Faster~\cite{Chen02}, SW-Faster~\cite{Saito03}, and CR-DA-DET~\cite{Xu06}) challenge this adaptation task, let alone source data-free. Even with such a wide range of the domain gap, it is surprising to see in Figure \ref{curve4} that the state-of-the-art methods are improved over a wide range of confidence thresholds. Especially when we use SED or SED+Mosaic, we can improve the mAP from 26.9\% of CR-DA-DET~\cite{Xu06} to 27.6\% and 29.0\%.

\subsubsection{Adaptation from Normal to Foggy Weather}
In real-world applications, object detectors may be used with different weather conditions. It is hard to collect and label a large number of data from every weather condition. To study the changing environment adaptation from normal weather to a foggy condition, Cityscapes and Foggy Cityscapes are used as the source domain and the target domain, respectively. The comparisons between our SFOD and other UDA object detection methods (i.e., DA-Faster~\cite{Chen02}, SW-Faster~\cite{Saito03}, DA-Detection~\cite{Hsu04}, CR-DA-DET~\cite{Xu06}, and AT-Faster~\cite{He05}) are presented on eight common categories in Table \ref{C2F}. Compared to 22.3\% mAP of the baseline, even using pseudo labels with label noise trained by SED and Mosaic can still be improved to 33.5\%. However, there is still a certain gap to achieve the performance of the traditional UDA object detection methods. As a further discussion, we used the same defogging method like DA-Detection~\cite{Hsu04} to improve the image quality of the target domain, and then studied the performance of SFOD under this condition. As we can see from Table \ref{C2F}, SFOD performance has been improved by approximately 3\% after defogging. Based on the above phenomenon, it can be concluded that the fog aggravates the label noise in pseudo labels, thus affecting the detection performance.

\subsection{Discussion and Analysis}
In SFOD, the training process with pseudo labels of target domain data obtained by using the source domain's pre-trained model will be disturbed because of noisy labels. Some object detection results are shown in Figure \ref{result}, whether using SED directly to search an appropriate confidence threshold for pseudo label generation or further combining with false negatives simulation, the negative effects brought by noisy labels can be well suppressed so that more objects can be detected. Our proposed SFOD achieves comparable even superior results to the existing source data based UDA methods, which means the source domain data is actually not fully exploited in the existing methods.

\section{Conclusion}
In this paper, we propose a new learning paradigm for unsupervised domain adaptive object detection named SFOD. The challenge lies in only utilizing a pre-trained model from the source domain instead of directly using source data to provide supervision signals. We make it solvable from the view of learning with noisy labels. Although our method even surpasses many source data-based methods, we have to admit that to completely remove noisy labels (false positives and false negatives) is still very difficult in an unsupervised way. This is a very critical problem in SFOD, and our work is the first try in this direction and hopes to bring more inspirations to the UDA community.

\bibliography{reference}

\end{document}